\documentclass[11pt]{article}

\usepackage[final]{acl}

\usepackage{times}
\usepackage{latexsym}
\usepackage[T1]{fontenc}
\usepackage[utf8]{inputenc}
\usepackage{microtype}
\usepackage{inconsolata}
\usepackage{graphicx}
\usepackage{booktabs}
\usepackage{multirow}
\usepackage{amsmath}
\usepackage{xcolor}

\usepackage{pgfplots}
\pgfplotsset{compat=1.18}
\definecolor{cGPT}{HTML}{1F77B4}
\definecolor{cQwen}{HTML}{D62728}
\definecolor{cCorrect}{HTML}{4C9A2A}
\definecolor{cSelect}{HTML}{F2A541}
\definecolor{cReason}{HTML}{8E6CBF}
\definecolor{cRetrieve}{HTML}{9E9E9E}
\usepackage{placeins}
\usepackage[most]{tcolorbox}
\usepackage{fvextra}
\usepackage{capt-of}
\usepackage{needspace}
\usepackage{CJKutf8}
\usepackage[capitalize]{cleveref}

\newcommand{\TBD}[1]{\if\relax\detokenize{#1}\relax\textcolor{red}{\textbf{?}}\else\textcolor{red}{\textbf{[TBD: #1]}}\fi}

\title{OpenTableRAG: Open-Domain Table Question Answering over Enterprise-Scale Database through Interactive Reasoning}

\author{
  \textbf{Yu-Hsuan Lin}\thanks{These authors contributed equally to this work.} \\
  National Taiwan University \\
  \texttt{r13922159@csie.ntu.edu.tw}
  \And
  \textbf{Tzu-Hsuan Chou}\footnotemark[1] \\
  CMoney Technology Corporation \\
  \texttt{kevin\_chou@cmoney.com.tw}
  \AND
  \textbf{Chun-Nan Chou} \\
  CMoney Technology Corporation \\
  \texttt{jason\_chou@cmoney.com.tw}
  \And
  \textbf{Yun-Nung Chen} \\
  National Taiwan University \\
  \texttt{y.v.chen@ieee.org}
}

\begin{document}
\maketitle

\begin{abstract}

The applicability of existing open-domain table question answering (OTQA) methods to large production databases remains uncertain.
We investigate OTQA in a financial data platform where full-table exports are prohibited, and SQL execution is resource-constrained.
We propose OpenTableRAG, extending TableRAG from closed-domain to open-domain table question answering under platform constraints.
Evaluated on real user queries over production tables, OpenTableRAG improves test-set accuracy over the Top-1 setting by more than $40$ percentage points with both GPT-6.1 Terra and Qwen3.8-27B.
The results demonstrate the feasibility of OTQA over enterprise-scale databases.

\end{abstract}

\section{Introduction}
\label{sec:intro}
 
In finance, data platforms commonly expose hundreds of tables covering market prices, financial statements, and institutional holdings.
Identifying relevant tables and formulating database queries can be challenging for users unfamiliar with the underlying database.
Open-domain table question answering (OTQA) systems have emerged as a promising solution to addressing these challenges by retrieving relevant tables from a corpus of tables and reasoning over table contents to answer natural-language questions~\citep{otqa}.

However, the applicability of existing OTQA methods to large production tables remains uncertain.
To the best of our knowledge, existing methods have been evaluated on datasets with substantially smaller tables.
OpenTab~\citep{kong2024opentab} and CRAFT~\citep{singh-etal-2026-craft} index full-table contents using BM25~\citep{robertson2009bm25} to evaluate on Open-WikiTable~\citep{kweon-etal-2023-open} and SPLADE~\citep{formal2021splade} to evaluate on NQ-Tables~\citep{herzig-etal-2021-open}, respectively.
As shown in Table~\ref{tab:corpus}, the median number of rows per table in our production corpus exceeds the corresponding medians of both evaluation datasets by more than four orders of magnitude.
The substantial difference in table size raises concerns about the scalability of full-table indexing in production environments.

Beyond table scale, our production financial data platform imposes two constraints on OTQA deployment.
First, the platform prohibits full-table exports.
As shown in Table~~\ref{tab:data-recency}, approximately half of the production tables are updated daily.
Indexes built from exported full-table contents may become outdated as the underlying data changes, potentially leading to incorrect answers.
Second, the platform imposes SQL execution constraints, including a $120$-second timeout and approximately $2$ GB of memory per API service instance.
Large SQL queries may thus fail due to execution timeouts or insufficient memory.
These platform constraints motivate us to develop an approach that retrieves current data through SQL queries and adapts query execution based on intermediate results and execution failures.

To address the challenges of large production tables and data access constraints, we propose OpenTableRAG, an OTQA pipeline that extends TableRAG~\citep{chen2024tablerag} from the closed-domain setting to the open-domain setting.
TableRAG is designed for question answering over large tables in the closed-domain setting, where the relevant table is given.
The multi-round reasoning mechanism in TableRAG provides a suitable foundation for handling the SQL execution constraints of our platform.
OpenTableRAG builds upon the foundation of TableRAG to support OTQA over enterprise-scale databases under platform constraints.
Our main contributions are fourfold:

\begin{itemize}
    \item We characterize a production financial table corpus, showing substantially larger table sizes than existing OTQA benchmarks and highlighting the recency of production data.
    \item We develop OpenTableRAG, extending TableRAG from closed-domain to the open-domain setting through metadata-based table retrieval, bounded cell indexing, and answer aggregation, without requiring full-table export.
    \item We evaluate OpenTableRAG on real user queries over our production tables, achieving accuracies of $78.29\%$ and $71.14\%$ with GPT-6.1 Terra and Qwen3.8-27B, respectively.
    \item We analyze end-to-end errors, identifying table retrieval as the primary error source and retrieval improvements as a promising direction for enhancing OpenTableRAG performance.
\end{itemize}

\begin{table}[t]
\centering
\small
\setlength{\tabcolsep}{3pt}
\begin{tabular}{@{}lrrr@{}}
\toprule
Dataset & NQ & OWT & Ours \\
\midrule
\#Tables & 169,898 & 24,680 & 578 \\
\#Questions & 11,628 & 67,023 & 280 \\
\addlinespace[2pt]
Median \#rows & 6 & 11 & 150,994 \\
Maximum \#rows & 2,185 & 1,950 & 254,383,645 \\
\addlinespace[2pt]
Median \#columns & 3 & 6 & 21 \\
Maximum \#columns & 430 & 44 & 645 \\
\bottomrule
\end{tabular}
\caption{
Comparison of table corpus statistics across NQ-Tables (NQ), Open-WikiTable (OWT), and our production table corpus (Ours).
}
\label{tab:corpus}
\end{table}

\begin{table}[t]
\centering
\small
\setlength{\tabcolsep}{8pt}
\begin{tabular}{@{}lrr@{}}
\toprule
Update window & \#Tables & \% \\
\midrule
Within 1 day   & 261 & 47.8 \\
Within 7 days  & 416 & 76.2 \\
Within 30 days & 473 & 86.6 \\
\bottomrule
\end{tabular}
\caption{
Recency of table updates in the production table corpus.
}
\label{tab:data-recency}
\end{table}

\section{OpenTableRAG: System Design}
\label{sec:design}

\begin{figure*}[t]
    \centering
    \includegraphics[width=2\columnwidth]{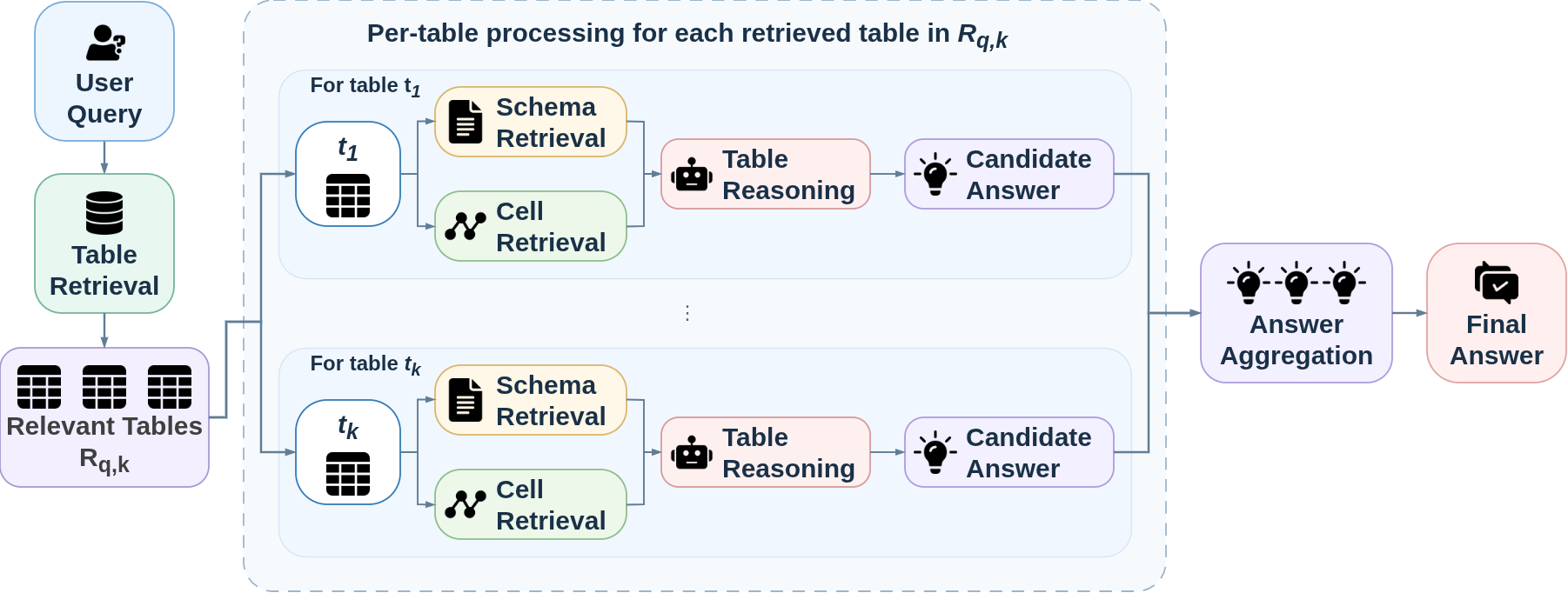}
    \caption{
        Overview of OpenTableRAG.
        Table retrieval identifies the top-$k$ relevant tables for a user query.
        Each retrieved table is processed independently through parallel schema and cell retrieval, followed by table reasoning.
        Answer aggregation selects the final answer from the per-table candidate answers.
    }
    \label{fig:pipeline}
\end{figure*}

As illustrated in Figure~\ref{fig:pipeline}, our proposed  OpenTableRAG consists of five stages: table retrieval, schema retrieval, cell retrieval, table reasoning, and answer aggregation.
The pipeline begins with table retrieval, which identifies relevant tables for a user query (Section~\ref{sec:table-retrieval}).
Each retrieved table is then processed independently through the three intermediate stages, which are built on TableRAG~\citep{chen2024tablerag}.
The schema retrieval and cell retrieval stages identify relevant columns and cell values in parallel (Sections~\ref{sec:schema-retrieval} and \ref{sec:cell-retrieval}).
The table reasoning stage then attempts to answer the query using the retrieved columns and cell values (Section~\ref{sec:table-reasoning}).
Finally, answer aggregation selects the final answer from the per-table answers (Section~\ref{sec:answer-aggregation}).
The following sections detail these five stages and the index update policy (Section~\ref{sec:index-update-policy}) designed to accommodate the data access constraints of our platform.

\subsection{Table Retrieval}
\label{sec:table-retrieval}

We construct a metadata-based index without full-table export, which is motivated by ReadSchema baseline of TableRAG~\citep{chen2024tablerag}.
Following the sparse and dense retrieval comparison in OpenTab~\citep{kong2024opentab}, we consider both approaches for table retrieval.
For each table in the production corpus, we index the table name and the associated column names, which are available on our platform.
Given a user query $q$, we retrieve the top-$k$ relevant tables, denoted by $R_{q,k}$.
Details of index construction and retrieval are provided in Appendix~\ref{app:table-retrieval}.

\subsection{Schema Retrieval}
\label{sec:schema-retrieval}

Schema retrieval identifies columns relevant to the user query $q$ within each retrieved table $t \in R_{q,k}$ without full-table export.
To increase coverage of relevant columns, a language model $L$ generates up to five\footnote{This hyperparameter for schema and cell retrieval is adopted directly from TableRAG.} expanded queries from $q$.
Retrieval is then performed using an index of column names extracted from table metadata on our platform.
Each expanded query is encoded to retrieve five columns from the index, and the union of retrieved columns contains at most $25$ distinct columns.
The query expansion and retrieval details are provided in Appendix~\ref{app:schema-retrieval}.

\subsection{Cell Retrieval}
\label{sec:cell-retrieval}

In parallel with schema retrieval, cell retrieval identifies column--value pairs relevant to the user query within each retrieved table.
Cell retrieval follows the same query expansion and retrieval procedures as schema retrieval.
However, indexing all distinct column--value pairs could require exporting large amounts of table content, conflicting data access constraints of our platform.
We therefore construct an index containing up to the $b$ most frequent column--value pairs across categorical columns.
The details of query expansion retrieval are provided in Appendix~\ref{app:cell-retrieval}.

\subsection{Table Reasoning}
\label{sec:table-reasoning}

Based on each retrieved table, table reasoning attempts to answer a user query.
The user query, together with the retrieved schema and cell information, is provided to a language model $L$ for reasoning.
Implementation details can be found in Appendix~\ref{app:table-reasoning}.

To accommodate the execution constraints of our platform, we adopt ReAct~\citep{yao2023react} for multi-round SQL querying.
Unlike Python-based solver~\citep{liu-etal-2024-rethinking} used by TableRAG, our implementation executes SQL queries generated by $L$ through the execution interface of our platform.
Query results and execution errors serve as observations, allowing $L$ to revise subsequent queries based on intermediate results and execution failures.

We allow at most $d$ ReAct iterations per retrieved table to limit reasoning cost.
An answer generated within $d$ iterations is returned as a candidate answer along with the corresponding SQL execution history.
Otherwise, table reasoning returns an unanswerable outcome.
Only candidate answers and the corresponding histories of SQL execution proceed to answer aggregation.

\subsection{Answer Aggregation}
\label{sec:answer-aggregation}

Answer aggregation selects a final answer from candidate answers produced by table reasoning.
We considers two strategies to select potentially conflicting candidate answers.
Implementation details of both strategies can be found in Appendix~\ref{app:answer-aggregation}.

\paragraph{SQL Similarity Ranking (SSR)}
Inspired by the SQL-based reranking strategy of OpenTab~\citep{kong2024opentab}, SSR ranks candidate answers using cosine similarity between embeddings of the user query $q$ and the final executed SQL of each candidate.
The candidate answer with the highest similarity becomes the final answer.

\paragraph{LLM-based Answer Selection (LAS)}
Inspired by the joint reasoning baseline of OpenTab~\citep{kong2024opentab}, LAS formulates answer aggregation as a classification task over independently generated candidate answers.
Given at most $k$ candidate answers, the language model $L$ either selects one candidate or rejects all candidates.
Unlike the joint reasoning baseline in OpenTab, LAS selects an answer from the results of independent table reasoning.
This design avoids combining information from all retrieved tables into a single LLM input, where relevant information may be overlooked.

\subsection{Index Update Policy}
\label{sec:index-update-policy}

To prevent stale indexes from affecting answer accuracy, we establish an index update policy.
Since table and column metadata rarely change, the corresponding table retrieval and schema retrieval indexes are updated once metadata changes occur.
In contrast, cell retrieval indexes require more frequent updates because approximately half of the production tables are updated daily, as shown in Table~\ref{tab:data-recency}.
Updating cell retrieval indexes involves recomputing the top-$b$ most frequent column--value pairs for each table and encoding the resulting documents.
We refresh all the indexes in cell retrieval daily at midnight to avoid periods of high traffic.

\section{Experiments}
\label{sec:experiments}

We evaluate OpenTableRAG on real user queries over our production financial table corpus.
We first describe the dataset and experimental settings, followed by evaluations of table retrieval and end-to-end OTQA performance.

\subsection{Dataset}
\label{sec:dataset}

Our dataset consists of $280$ natural-language queries over a production corpus (Table~\ref{tab:corpus}), split evenly between development and test sets, i.e., each contains $140$ queries.
The queries were collected from real users.
Each query can be answered using a single table in the corpus.
Some queries can be answered independently using different tables, yielding the same answer.
All reference answers were manually written and verified.

\subsection{Settings}
\label{sec:settings}

\paragraph{Hyperparameter Selection}
We select the hyperparameters of our proposed OpenTableRAG based on the development-set results.
We set the number of retrieved tables $k$ to $15$, the limit of cell index size $b$ to $5{,}000$, and the maximum number of ReAct iterations $d$ to $12$.
The justification for the hyperparameter choices is provided in Appendix~\ref{app:hyperparameters}.

\paragraph{Language Models}
We evaluate two language models for $L$: GPT-6.1 Terra~\citep{openai2026terra} and Qwen3.8-27B~\citep{qwen38}.
We select GPT-6.1 Terra based on expected high performance and acceptable experimental cost.
We choose Qwen3.8-27B as the most capable open-weight model we could deploy on our in-house GPU servers.
Within each run, we utilize the same $L$ in query expansion, table reasoning, and answer aggregation stages.
The details of deployment and inference settings are provided in Appendix~\ref{app:language-model-settings}.

\paragraph{Comparisons}
We compare OpenTableRAG with two settings to evaluate the impact of table identification and answer aggregation.
The \textbf{top-1} setting uses only the highest-ranked retrieved table to evaluate whether single-table retrieval is sufficient for OTQA.
The \textbf{closed-domain} setting uses a ground-truth table to evaluate question-answering performance when the correct table is known.
Both settings use the same schema retrieval, cell retrieval, and table reasoning procedures as those in our OpenTableRAG.
Neither the top-1 nor closed-domain require answer aggregation because only one table is used for each query in these two settings.

\paragraph{Evaluation}
We evaluate the performance of table retrieval and end-to-end question answering using Hit@$k$ and accuracy, respectively.
For each query, Hit@$k$ is 1 if at least one ground-truth table is included in the top-$k$ retrieved tables, and 0 otherwise.
We average Hit@$k$ across all user queries.
Accuracy is determined by rule-based matching between predicted and reference answers.
We report the mean and standard deviation of test-set accuracy across five runs.
The answer evaluation rules are provided in Appendix~\ref{app:evaluation-rules}.

\subsection{Table Retrieval Performance}
\label{sec:exp-table-retrieval}

Table retrieval is effective without full-table exports.
As illustrated in Table~\ref{tab:table-retrieval}, both sparse and dense retrieval achieve Hit@$15$ of at least 50\% on the development set using table metadata alone.
Dense retrieval outperforms sparse retrieval across all evaluated values of $k$.
Based on the development-set results in Table~\ref{tab:table-retrieval}, we adopt dense retrieval as our table retrieval method for all subsequent experiments.
The test-set results are reported for evaluation purposes only and are not used for the selection of retrieval methods.

\begin{table}[t]
\centering
\small
\setlength{\tabcolsep}{4pt}
\begin{tabular}{@{}llcccc@{}}
\toprule
Split & Method & H@1 & H@5 & H@10 & H@15 \\
\midrule
\multirow{2}{*}{Dev}  & Sparse & 22.14 & 37.86 & 44.29 & 50.00 \\
                      & Dense  & \textbf{35.00} & \textbf{65.00} & \textbf{73.57} & \textbf{78.57} \\
\midrule
\multirow{2}{*}{Test} & Sparse & 26.43 & 45.71 & 54.29 & 61.43 \\
                      & Dense  & \textbf{33.57} & \textbf{65.00} & \textbf{78.57} & \textbf{80.00} \\
\bottomrule
\end{tabular}
\caption{Table retrieval Hit@$k$ (\%) on the development and test sets, abbreviated as H@$k$ in the table.}
\label{tab:table-retrieval}
\end{table}

\subsection{OTQA Performance}
\label{sec:exp-otqa}

Answer aggregation could improves OTQA performance.
As shown in Table~\ref{tab:main}, both SSR and LAS achieve higher test-set accuracy than the Top-1 setting for both language models.
The highest-ranked retrieved table often fails to answer the query, whereas answer aggregation can select a correct answer from lower-ranked tables.

LAS outperforms SSR.
As illustrated in Table~\ref{tab:main}, LAS improves test-set accuracy over SSR by more than 30 percentage points for both language models.
Inspection of SSR results suggests that SQL queries generated for irrelevant tables may still have high embedding similarity with the user query, leading to incorrect answer selection.
In contrast, LAS selects an answer from candidate reasoning results rather than relying on SQL embedding similarity, potentially explaining its higher accuracy.


Retrieval and aggregation remain challenging.
As illustrated in Table~\ref{tab:main}, the performance gap between OpenTableRAG with LAS and the closed-domain setting is nearly $20$ percentage points for both language models.
The closed-domain results demonstrate the combined effectiveness of schema retrieval, cell retrieval, and table reasoning when the ground-truth table is known.
The remaining gap suggests room for improvement in table retrieval and answer aggregation.

\begin{table}[t]
\centering
\small
\setlength{\tabcolsep}{4pt}
\begin{tabular}{@{}lcc@{}}
\toprule
Setting & GPT-6.1 Terra & Qwen3.8-27B \\
\midrule
\multicolumn{3}{@{}l}{\textit{Open-domain}} \\
Top-1
& 31.71$_{\pm 0.39}$
& 29.29$_{\pm 1.52}$ \\
OpenTableRAG + SSR
& 36.00$_{\pm 3.37}$
& 39.86$_{\pm 3.89}$ \\
OpenTableRAG + LAS
& $\mathbf{78.29}_{\pm 0.39}$
& $\mathbf{71.14}_{\pm 1.20}$ \\
\midrule
Closed-domain
& $\mathit{96.86}_{\pm 0.64}$
& $\mathit{90.57}_{\pm 1.55}$ \\
\bottomrule
\end{tabular}
\caption{
Test-set accuracy (\%), with standard deviations shown as subscripts.
}
\label{tab:main}
\end{table}

\section{System Analysis Under Platform Constraints}
\label{sec:analysis}

We further analyze OpenTableRAG under the constraints of our platform.
We examine SQL error recovery through multi-round querying, performance on large production tables, and the distribution of errors across different stages.

\subsection{Multi-Round SQL Querying}
\label{sec:analysis-multi-round-sql-querying}

We evaluate the capability of multi-round SQL querying to recover from SQL errors under the execution constraints of our production platform.
To analyze SQL errors independently of the other stages in OpenTableRAG, we conduct the evaluation in the closed-domain setting on the test set.

To evaluate the recovery capability for SQL errors, we define three metrics.
First, Incidence measures the proportion of user queries encountering at least one SQL error during table reasoning.
Among queries encountering SQL errors, Recovery measures the proportion that successfully execute a subsequent SQL query, while Success measures the proportion that recover and produce a correct final answer.
We report the mean and standard deviation across five runs for each metric.

Multi-round SQL querying supports recovery under execution constraints.
As illustrated in Table~\ref{tab:sql-error-recovery}, the low Incidence of GPT-6.1 Terra leaves few SQL error cases.
As a result, Recovery and Success exhibit high standard deviations, limiting the reliability of recovery evaluation.
In contrast, Qwen3.8-27B encounters SQL errors in approximately $38\%$ of queries but achieves Recovery above $90\%$ and Success above $80\%$.
These results suggest that the proposed multi-round SQL querying mechanism can effectively recover from SQL errors and produce correct answers.
An illustrative example of SQL revision following a timeout is provided in Figure~\ref{fig:prompt-table-reasoning} in the Appendix.

\begin{table}[t]
\centering
\small
\begin{tabular}{lcc}
\toprule
Metric & GPT-6.1 Terra & Qwen3.8-27B \\
\midrule
Incidence
    & $1.61_{\pm 0.66}$
    & $38.17_{\pm 1.32}$ \\
Recovery
    & $86.00_{\pm 21.91}$
    & $94.67_{\pm 2.05}$ \\
Success
    & $38.00_{\pm 24.68}$
    & $84.24_{\pm 2.60}$ \\
\bottomrule
\end{tabular}

\caption{
Incidence, Recovery, and Success (\%) in closed-domain table reasoning on the test set, with standard deviations shown as subscripts.
}

\label{tab:sql-error-recovery}
\end{table}

\subsection{Large Production Tables}
\label{sec:large-production-tables}

OpenTableRAG supports question answering over large production tables, although accuracy remains lower for queries involving larger tables.
We divide test-set queries into two groups based on the median table size of $150,994$ rows in our production corpus (Table~\ref{tab:corpus}).
This threshold substantially exceeds the maximum table sizes in NQ-Tables ($2,185$ rows) and Open-WikiTable ($1,950$ rows).
As shown in Table~6, OpenTableRAG with LAS achieves over $60\%$ accuracy on the larger-table group for both language models.
However, accuracy is lower for the larger-table group than for the smaller-table group.
These results highlight the feasibility of OTQA at scales beyond existing benchmarks while indicating the need for further improvements on larger production tables.

\begin{table}[t]
\centering
\small
\setlength{\tabcolsep}{4pt}
\begin{tabular}{@{}lccc@{}}
\toprule
\#Rows & \#Queries & GPT-6.1 Terra & Qwen3.8-27B \\
\midrule
$\leq$ 150,994 & 56 & 88.57$_{\pm 0.98}$ & 81.79$_{\pm 1.49}$ \\
$>$ 150,994    & 84 & 71.43$_{\pm 0.00}$ & 64.05$_{\pm 1.55}$ \\
\bottomrule
\end{tabular}

\caption{
Accuracy (\%) of OpenTableRAG with LAS on the test set, grouped by the number of rows in the ground-truth table.
The threshold of $150,994$ rows is the median number of rows per table in our production corpus (Table~\ref{tab:corpus}).
Results are averaged over five runs, with standard deviations shown as subscripts.
}

\label{tab:table-size}
\end{table}

\subsection{Error Decomposition}
\label{sec:analysis-error-decomposition}

Schema and cell retrieval are unlikely to be the primary sources of errors when the correct table is given.
As shown in Table~\ref{tab:schema-cell}, schema and cell retrieval achieve hit rates of approximately $98\%$ and $85\%$, respectively, in the closed-domain setting on the test set.
A hit is defined as retrieving at least one gold column or gold column--value pair for schema and cell retrieval, respectively.
Missing gold items does not necessarily lead to incorrect answers, as language models may still generate correct SQL queries.
We therefore focus the subsequent error analysis on table retrieval, table reasoning, and answer aggregation.

\begin{figure}[t]
\centering
\includegraphics[width=\columnwidth]{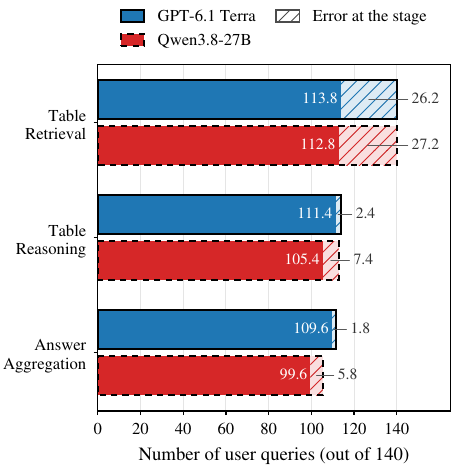}

\caption{
Error decomposition of OpenTableRAG with LAS on the test set, averaged over five runs.
Each bar represents the queries entering a stage.
The solid portion shows queries passing the stage, while the hatched portion shows queries failing at that stage.
Table retrieval errors occur when no ground-truth table is retrieved and no correct candidate answer is generated.
Table reasoning errors occur when a ground-truth table is retrieved but no correct candidate answer is generated.
Answer aggregation errors occur when a correct candidate answer is generated but the final answer is incorrect.
}

\label{fig:error-decomposition}
\end{figure}

Table retrieval is the primary source of end-to-end errors.
As shown in Figure~\ref{fig:error-decomposition}, table retrieval accounts for the majority of errors for both language models in OpenTableRAG with LAS.
Fewer errors occur during table reasoning and answer aggregation, although Qwen3.8-27B encounters more errors than GPT-6.1 Terra in both stages.
These results suggest that improving table retrieval is a promising direction for enhancing OpenTableRAG performance.

\begin{table}[t]
\centering
\small
\setlength{\tabcolsep}{4pt}
\begin{tabular}{@{}lcc@{}}
\toprule
Stage & GPT-6.1 Terra & Qwen3.8-27B \\
\midrule
Schema & 98.43$_{\pm 0.32}$ & 98.00$_{\pm 0.78}$ \\
Cell & 85.51$_{\pm 0.43}$ & 85.35$_{\pm 1.20}$ \\
\bottomrule
\end{tabular}
\caption{
Schema and cell retrieval hit rates in the closed-domain setting on the test set.
Results are reported as percentages, with means across five runs and standard deviations shown as subscripts.
}
\label{tab:schema-cell}
\end{table}

\section{Conclusion}
\label{sec:conclusion}

We developed OpenTableRAG to address the challenges of large production tables under platform constraints.
Extending TableRAG to the open-domain setting, OpenTableRAG combines metadata-based table retrieval, bounded cell indexing, and answer aggregation without requiring full-table exports.
We adapt ReAct in OpenTableRAG for multi-round SQL querying, allowing query revision based on execution results and errors.
Experiments on real user queries over production tables show that OpenTableRAG outperforms the top-1 setting with both GPT-6.1 Terra and Qwen3.8-27B.
System analyses further show that multi-round SQL querying supports recovery from SQL errors, while table retrieval is the primary source of end-to-end errors.
OpenTableRAG demonstrates the feasibility of OTQA over production tables substantially larger than existing OTQA benchmarks, although accuracy remains lower for queries involving larger production tables.
The experimental findings highlight improving table retrieval and question answering over larger production tables as directions for future research.

\section*{Limitations}

Our evaluation is conducted on a single production financial data platform.
Generalizability to other domains, database systems, and languages remains unverified.
Although individual tables in our production corpus can contain hundreds of millions of rows, the corpus contains only 578 tables.
The scalability of table retrieval to substantially larger table collections remains unverified.
Although our pipeline avoids full-table export, reasoning independently over multiple retrieved tables may incur high SQL execution costs.
We do not provide a comprehensive end-to-end efficiency comparison with existing OTQA methods.

\bibliography{custom}

@inproceedings{otqa,
    title = "Open Domain Question Answering over Tables via Dense Retrieval",
    author = {Herzig, Jonathan  and
      M{\"u}ller, Thomas  and
      Krichene, Syrine  and
      Eisenschlos, Julian},
    editor = "Toutanova, Kristina  and
      Rumshisky, Anna  and
      Zettlemoyer, Luke  and
      Hakkani-Tur, Dilek  and
      Beltagy, Iz  and
      Bethard, Steven  and
      Cotterell, Ryan  and
      Chakraborty, Tanmoy  and
      Zhou, Yichao",
    booktitle = "Proceedings of the 2021 Conference of the {N}orth {A}merican Chapter of the {A}ssociation for {C}omputational {L}inguistics: Human Language Technologies",
    month = jun,
    year = "2021",
    address = "Online",
    publisher = "Association for Computational Linguistics",
    url = "https://aclanthology.org/2021.naacl-main.43/",
    doi = "10.18653/v1/2021.naacl-main.43",
    pages = "512--519"
}

@inproceedings{kong2024opentab,
	title        = {OpenTab: Advancing Large Language Models as Open-domain Table Reasoners},
	author       = {Kezhi Kong and Jiani Zhang and Zhengyuan Shen and Balasubramaniam Srinivasan and Chuan Lei and Christos Faloutsos and Huzefa Rangwala and George Karypis},
	year         = 2024,
	booktitle    = {The Twelfth International Conference on Learning Representations},
	url          = {https://openreview.net/forum?id=Qa0ULgosc9}
}

@inproceedings{singh-etal-2026-craft,
	title        = {{CRAFT}: Training-Free Cascaded Retrieval for Tabular {QA}},
	author       = {Singh, Adarsh  and Bhandari, Kushal Raj  and Gao, Jianxi  and Dan, Soham  and Gupta, Vivek},
	year         = 2026,
	month        = jul,
	booktitle    = {Proceedings of the 64th Annual Meeting of the {A}ssociation for {C}omputational {L}inguistics (Volume 1: Long Papers)},
	publisher    = {Association for Computational Linguistics},
	address      = {San Diego, California, United States},
	pages        = {3284--3298},
	doi          = {10.18653/v1/2026.acl-long.149},
	isbn         = {979-8-89176-390-6},
	url          = {https://aclanthology.org/2026.acl-long.149/},
	editor       = {Liakata, Maria  and Moreira, Viviane P.  and Zhang, Jiajun  and Jurgens, David}
}

@inproceedings{kweon-etal-2023-open,
	title        = {Open-{W}iki{T}able : Dataset for Open Domain Question Answering with Complex Reasoning over Table},
	author       = {Kweon, Sunjun  and Kwon, Yeonsu  and Cho, Seonhee  and Jo, Yohan  and Choi, Edward},
	year         = 2023,
	month        = jul,
	booktitle    = {Findings of the {A}ssociation for {C}omputational {L}inguistics: {ACL} 2023},
	publisher    = {Association for Computational Linguistics},
	address      = {Toronto, Canada},
	pages        = {8285--8297},
	doi          = {10.18653/v1/2023.findings-acl.526},
	url          = {https://aclanthology.org/2023.findings-acl.526/},
	editor       = {Rogers, Anna  and Boyd-Graber, Jordan  and Okazaki, Naoaki}
}

@inproceedings{herzig-etal-2021-open,
	title        = {Open Domain Question Answering over Tables via Dense Retrieval},
	author       = {Herzig, Jonathan  and M{\"u}ller, Thomas  and Krichene, Syrine  and Eisenschlos, Julian},
	year         = 2021,
	month        = jun,
	booktitle    = {Proceedings of the 2021 Conference of the {N}orth {A}merican Chapter of the {A}ssociation for {C}omputational {L}inguistics: Human Language Technologies},
	publisher    = {Association for Computational Linguistics},
	address      = {Online},
	pages        = {512--519},
	doi          = {10.18653/v1/2021.naacl-main.43},
	url          = {https://aclanthology.org/2021.naacl-main.43/},
	editor       = {Toutanova, Kristina  and Rumshisky, Anna  and Zettlemoyer, Luke  and Hakkani-Tur, Dilek  and Beltagy, Iz  and Bethard, Steven  and Cotterell, Ryan  and Chakraborty, Tanmoy  and Zhou, Yichao}
}

@inproceedings{chen2024tablerag,
	title        = {TableRAG: Million-Token Table Understanding with Language Models},
	author       = {Chen, Si-An and Miculicich, Lesly and Eisenschlos, Julian Martin and Wang, Zifeng and Wang, Zilong and Chen, Yanfei and Fujii, Yasuhisa and Lin, Hsuan-Tien and Lee, Chen-Yu and Pfister, Tomas},
	year         = 2024,
	booktitle    = {Advances in Neural Information Processing Systems},
	publisher    = {Curran Associates, Inc.},
	volume       = 37,
	pages        = {74899--74921},
	doi          = {10.52202/079017-2382},
	url          = {https://proceedings.neurips.cc/paper_files/paper/2024/file/88dd7aa6979e352fda7c4952ca8eac59-Paper-Conference.pdf},
	editor       = {A. Globerson and L. Mackey and D. Belgrave and A. Fan and U. Paquet and J. Tomczak and C. Zhang}
}

@article{robertson2009bm25,
	title        = {The Probabilistic Relevance Framework: BM25 and Beyond},
	author       = {Robertson, Stephen and Zaragoza, Hugo},
	year         = 2009,
	month        = apr,
	journal      = {Found. Trends Inf. Retr.},
	publisher    = {Now Publishers Inc.},
	address      = {Hanover, MA, USA},
	volume       = 3,
	number       = 4,
	pages        = {333–389},
	doi          = {10.1561/1500000019},
	issn         = {1554-0669},
	url          = {https://doi.org/10.1561/1500000019},
	issue_date   = {April 2009},
	numpages     = 57
}

@inproceedings{formal2021splade,
	title        = {SPLADE: Sparse Lexical and Expansion Model for First Stage Ranking},
	author       = {Formal, Thibault and Piwowarski, Benjamin and Clinchant, St{\'e}phane},
	year         = 2021,
	booktitle    = {Proceedings of the 44th International ACM SIGIR Conference on Research and Development in Information Retrieval},
	location     = {Virtual Event, Canada},
	publisher    = {Association for Computing Machinery},
	address      = {New York, NY, USA},
	series       = {SIGIR '21},
	pages        = {2288–2292},
	doi          = {10.1145/3404835.3463098},
	isbn         = 9781450380379,
	url          = {https://doi.org/10.1145/3404835.3463098},
	numpages     = 5
}

@inproceedings{yao2023react,
	title        = {ReAct: Synergizing Reasoning and Acting in Language Models},
	author       = {Shunyu Yao and Jeffrey Zhao and Dian Yu and Nan Du and Izhak Shafran and Karthik R Narasimhan and Yuan Cao},
	year         = 2023,
	booktitle    = {The Eleventh International Conference on Learning Representations},
	url          = {https://openreview.net/forum?id=WE_vluYUL-X}
}

@inproceedings{liu-etal-2024-rethinking,
    title = "Rethinking Tabular Data Understanding with Large Language Models",
    author = "Liu, Tianyang  and
      Wang, Fei  and
      Chen, Muhao",
    editor = "Duh, Kevin  and
      Gomez, Helena  and
      Bethard, Steven",
    booktitle = "Proceedings of the 2024 Conference of the {N}orth {A}merican Chapter of the {A}ssociation for {C}omputational {L}inguistics: Human Language Technologies (Volume 1: Long Papers)",
    month = jun,
    year = "2024",
    address = "Mexico City, Mexico",
    publisher = "Association for Computational Linguistics",
    url = "https://aclanthology.org/2024.naacl-long.26/",
    doi = "10.18653/v1/2024.naacl-long.26",
    pages = "450--482"
}

@inproceedings{gunther-etal-2025-jina,
    title = "jina-embeddings-v4: Universal Embeddings for Multimodal Multilingual Retrieval",
    author = {G{\"u}nther, Michael  and
      Sturua, Saba  and
      Akram, Mohammad Kalim  and
      Mohr, Isabelle  and
      Ungureanu, Andrei  and
      Wang, Bo  and
      Eslami, Sedigheh  and
      Martens, Scott  and
      Werk, Maximilian  and
      Wang, Nan  and
      Xiao, Han},
    editor = "Adelani, David Ifeoluwa  and
      Arnett, Catherine  and
      Ataman, Duygu  and
      Chang, Tyler A.  and
      Gonen, Hila  and
      Raja, Rahul  and
      Schmidt, Fabian  and
      Stap, David  and
      Wang, Jiayi",
    booktitle = "Proceedings of the 5th Workshop on Multilingual Representation Learning ({MRL} 2025)",
    month = nov,
    year = "2025",
    address = "Suzhuo, China",
    publisher = "Association for Computational Linguistics",
    url = "https://aclanthology.org/2025.mrl-main.36/",
    doi = "10.18653/v1/2025.mrl-main.36",
    pages = "531--550",
    ISBN = "979-8-89176-345-6"
}

@misc{qwen38,
    title = {{Qwen3.8-Max}: A New Bar for Coding and Cowork},
    url = {https://qwen.ai/blog?id=qwen3.8},
    author = {{Qwen Team}},
    month = {August},
    year = {2026}
}

@inproceedings{kwon2023efficient,
  title={Efficient Memory Management for Large Language Model Serving with PagedAttention},
  author={Woosuk Kwon and Zhuohan Li and Siyuan Zhuang and Ying Sheng and Lianmin Zheng and Cody Hao Yu and Joseph E. Gonzalez and Hao Zhang and Ion Stoica},
  booktitle={Proceedings of the ACM SIGOPS 29th Symposium on Operating Systems Principles},
  year={2023}
}

@techreport{openai2026terra,
  author      = {{OpenAI}},
  title       = {Addendum to {GPT}-6 {A}stra System Card: {GPT}-6.1 {T}erra},
  institution = {OpenAI},
  year        = {2026},
  url         = {https://openai.com},
  note        = {Accessed: 2026-11-01}
}

\appendix

\section{Hyperparameter Selection}
\label{app:hyperparameters}

We justify the choices
for the number of retrieved tables $k$, the limit of cell index size $b$, and the maximum number of ReAct iterations $d$.

\subsection{Number of Retrieved Tables}
\label{app:exp-table-retrieval}

\begin{figure}[!t]
\centering
\includegraphics[width=\columnwidth]{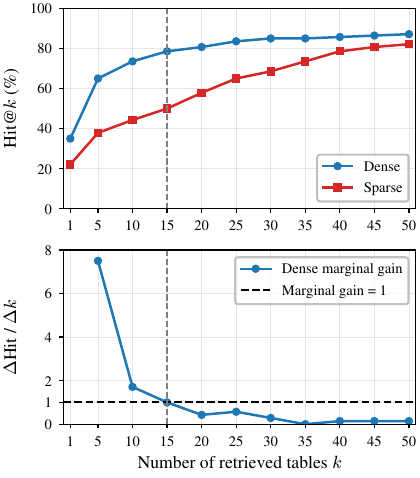}

\caption{
Hit@$k$ and marginal gain for table retrieval on the development set.
The top plot compares sparse and dense retrieval, while the bottom plot shows the marginal gain of Hit@$k$ for dense retrieval.
The vertical dashed lines indicate the selected $k=15$, and the horizontal dashed line indicates a marginal gain of one percentage point per additional retrieved table.
}

\label{fig:table-retrieval}
\end{figure}

We select the number of retrieved tables $k$ based on the trade-off between retrieval effectiveness and downstream costs.
As illustrated in the top plot of Figure~\ref{fig:table-retrieval}, increasing $k$ improves Hit@$k$ for both sparse and dense retrieval methods.
However, retrieving more tables may introduce irrelevant candidates and increase the token usage of table reasoning and answer aggregation.
We consequently evaluate the trade-off across selected values of $k$ ranging from $1$ to $50$.
The experiments use sparse and dense retrieval on the development set.

To quantify the benefit of retrieving additional tables, we calculate the marginal gain of Hit@$k$.
The marginal gain measures the change in Hit@$k$ divided by the change in $k$ between consecutive evaluated settings.
The marginal gain is expressed in percentage points per additional retrieved table.

The marginal gain generally decreases as $k$ increases.
As illustrated in the bottom plot of Figure~\ref{fig:table-retrieval}, the marginal gain falls below one percentage point per additional retrieved table for evaluated values of $k$ greater than $15$.
We therefore set the number of retrieved tables $k$ to $15$ for all other experiments.

\subsection{Cell Index Size Limit}
\label{app:exp-cell-budget}

We select the cell index size limit $b$ based on the trade-off between cell retrieval effectiveness and computational cost.
Increasing $b$ can make more query-relevant column--value pairs available for retrieval.
However, a larger cell index increases index embedding time and retrieval latency.
We consequently evaluate the trade-off across selected values of $b$ ranging from $1{,}000$ to $100{,}000$.
The experiments use the gold tables of the development set and Qwen3.8-27B~\citep{qwen38} for query expansion.

We evaluate cell retrieval effectiveness using average recall and hit.
Recall measures the proportion of gold column--value pairs retrieved for each query, while hit measures the proportion of queries for which at least one gold pair is retrieved.
We also measure coverage as the proportion of gold column--value pairs included in the cell index.
We evaluate computational cost using index embedding time and retrieval latency.
Recall, hit, and per-query retrieval latency are averaged over five runs.
Index embedding time is measured across all gold tables, while retrieval latency includes keyword encoding and exact nearest-neighbor search.

\begin{figure}[t]
\centering
\includegraphics[width=\columnwidth]{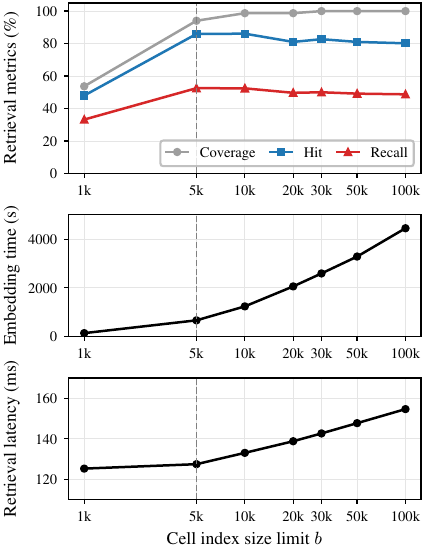}
\caption{
Effect of the cell index size limit $b$ on coverage, average recall, and hit (top panel), index embedding time (middle panel), and retrieval latency (bottom panel).
The horizontal axis uses a logarithmic scale.
The vertical dashed lines indicate the selected $b=5{,}000$.
}
\label{fig:cell-budget}
\end{figure}

Increasing $b$ improves coverage but does not necessarily improve cell retrieval effectiveness.
Figure~\ref{fig:cell-budget} shows that coverage increases with $b$, while average recall peaks at $b=5{,}000$ and generally declines at larger index sizes.
Hit reaches its maximum at $b=10{,}000$, only 0.17 percentage points higher than at $b=5{,}000$.
The marginal improvement suggests limited benefit from further enlarging the cell index.
On the cost side, retrieval latency remains relatively stable across the evaluated values of $b$.
However, increasing $b$ from $5{,}000$ to $10{,}000$ raises index embedding time by 86.0\%.
We therefore set the cell index size limit $b$ to $5{,}000$ for all other experiments.

\subsection{Number of ReAct Iterations}
\label{app:exp-react-iterations}

We assess the maximum number of ReAct iterations $d$ based on the trade-off between table reasoning accuracy and token usage.
Additional iterations allow more SQL query revisions, potentially improving reasoning accuracy. 
However, more iterations may increase token usage.
We run table reasoning for up to $d=15$ iterations.

We evaluate table reasoning accuracy in a closed-domain setting on the development set.
A ground-truth table is provided for each query.
Schema retrieval and cell retrieval provide the retrieved columns and column--value pairs, as in OpenTableRAG.
For smaller values of $d$, we evaluate whether a correct final answer has been produced within the first $d$ iterations.
We omit $d=1$ because the first iteration always issues an SQL query, resulting in zero accuracy.
Accuracy is averaged over five independent runs.

\begin{figure}[t]
\centering
\includegraphics[width=\columnwidth]{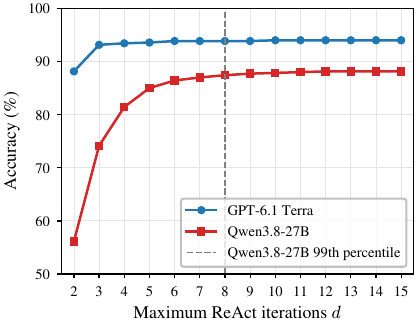}
\caption{
Closed-domain table reasoning accuracy (\%) across different maximum numbers of ReAct iterations $d$ on the development set.
The vertical dashed line indicates the 99th percentile of iterations required by Qwen3.8-27B to correctly answer user queries, which is used to select $d$.
}
\label{fig:react-iterations}
\end{figure}

Additional ReAct iterations benefit Qwen3.8-27B more than GPT-6.1 Terra.
Figure~\ref{fig:react-iterations} shows that GPT-6.1 Terra approaches the highest observed accuracy within three iterations, whereas Qwen3.8-27B continues to improve with additional iterations.
To determine a shared iteration limit, we calculate the 99th percentile of iterations required by Qwen3.8-27B to correctly answer user queries.
We multiply the 99th-percentile iteration count of 8 by 1.5 to offer a sufficient reasoning budget.
We therefore set the maximum number of ReAct iterations $d$ to $12$ for all other experiments.

\section{Implementation Details}
\label{app:details}

\subsection{Dense Retrieval Settings}
\label{app:dense-retrieval-settings}

To perform dense retrieval, we calculate document embeddings using a document encoder $\mathcal{E}_{\mathrm{doc}}$ and construct an index of the resulting embeddings.
Retrieval queries are encoded using a query encoder $\mathcal{E}_{\mathrm{query}}$.
For a retrieval query $x$ and a document $y$, the relevance score is computed using cosine similarity between query and document embeddings:
\begin{equation}
    r(x,y) =
    \frac{
        \langle \mathcal{E}_{\mathrm{query}}(x), \mathcal{E}_{\mathrm{doc}}(y) \rangle
    }{
        \|\mathcal{E}_{\mathrm{query}}(x)\|
        \|\mathcal{E}_{\mathrm{doc}}(y)\|
    },
\end{equation}
where $\langle \cdot, \cdot \rangle$ and $\|\cdot\|$ denote inner product and Euclidean norm, respectively.
All dense retrieval stages use exact nearest-neighbor search based on cosine similarity.

The same embedding model and encoder configurations are used across all retrieval stages, while the documents and queries vary across stages, as specified in the corresponding subsections.
We use the Hugging Face Transformers API to load \texttt{jina-embeddings-v4}\footnote{\url{https://huggingface.co/jinaai/jina-embeddings-v4}}~\citep{gunther-etal-2025-jina} and generate embeddings using the \texttt{encode\_text} method with \texttt{task=retrieval}.
For $\mathcal{E}_{\mathrm{doc}}$, we set \texttt{prompt\_name=passage}, which automatically prepends \texttt{"Passage:"} during encoding.
For $\mathcal{E}_{\mathrm{query}}$, we set \texttt{prompt\_name=query}, which automatically prepends \texttt{"Query:"} during encoding.

\subsection{Language Model Settings}
\label{app:language-model-settings}

We deploy GPT-6.1 Terra~\citep{openai2026terra} and Qwen3.8-27B~\citep{qwen38} through different serving environments.
GPT-6.1 Terra is accessed through Azure OpenAI\footnote{\url{https://azure.microsoft.com/}}.
Qwen3.8-27B is deployed on our in-house GPU servers using vLLM~\citep{kwon2023efficient} to provide a consistent API interface across both models.

We use the same sampling parameters for both models.
Following TableRAG~\citep{chen2024tablerag}, we set temperature to $0.8$ and top-$p$ to $0.95$.
We limit the maximum output length to $4,096$ tokens, except for query expansion, which is limited to $512$ tokens.
For GPT-6.1 Terra, we set the reasoning effort to none.

\subsection{Table Retrieval}
\label{app:table-retrieval}

We construct a metadata-based index for the production table corpus $T$.
For each table $t \in T$, we concatenate its table name and column names into a single document $s_t$.
As the table and column names are in Traditional Chinese, we use the Chinese enumeration comma (Unicode U+3001) to join the names.

For sparse retrieval, we construct a BM25~\citep{robertson2009bm25} index over the documents $s_t$.
We use the original user query $q$ to rank tables by BM25 relevance scores.

For dense retrieval, each document is encoded using $\mathcal{E}_{\mathrm{doc}}$ to construct the embedding index.
Table retrieval ranks each table $t \in T$ by relevance score $r(q, s_t)$ to the user query $q$.
The top-$k$ retrieved tables are defined as:
\begin{equation}
    R_{q,k}
    =
    \operatorname{Topk}_{t \in T}
    r(q,s_t),
\end{equation}
where $\operatorname{Topk}$ returns the $k$ highest-scoring tables.

\subsection{Schema Retrieval}
\label{app:schema-retrieval}

We construct one schema retrieval index for each table $t \in T$.
Let $C_t$ denote the set of column names of table $t$.
We encode each column name $c \in C_t$ using $\mathcal{E}_{\mathrm{doc}}$.

For each retrieved table $t \in R_{q,k}$, schema retrieval expands a given user query and retrieves relevant columns.
A language model $L$ generates up to five expanded queries $Q_{q,t}^{\mathrm{schema}}$ from the user query $q$.
Each expanded query $\hat{q} \in Q_{q,t}^{\mathrm{schema}}$ is encoded using $\mathcal{E}_{\mathrm{query}}$ to rank column names by the relevance score $r(\hat{q},c)$.
We take the union of the top-five columns retrieved for each expanded query:
\begin{equation}
    S_{q,t}
    =
    \bigcup_{\hat{q} \in Q_{q,t}^{\mathrm{schema}}}
    \operatorname{Top5}_{c \in C_t}
    r(\hat{q},c),
\end{equation}
where $\operatorname{Top5}$ returns five columns with the highest relevance scores.
The schema query expansion prompt is provided in Appendix~\ref{app:schema-query-expansion-prompts}.

We enrich retrieved columns with data types and representative values for table reasoning.
Data types are obtained from metadata on our platform, while representative values are precomputed and stored using separate SQL aggregation queries for each column.
We retrieve minimum and maximum values for numerical columns and the three most frequent values for categorical columns.
If an aggregation query fails due to execution limits, we omit the corresponding representative values but retain the column name and data type.

\subsection{Cell Retrieval}
\label{app:cell-retrieval}

We construct one cell retrieval index for each table $t \in T$.
Using separate SQL aggregation queries for each categorical column, we obtain up to $b$ most frequent values and the corresponding frequencies.
If an aggregation query fails due to execution limits, we omit values from the affected column while retaining results from successful queries.
We then select the $b$ most frequent column--value pairs across columns and serialize each pair as a JSON document, forming the document set $P_t^{(b)}$.
Each document is encoded using $\mathcal{E}_{\mathrm{doc}}$ and follows the format:

\begin{verbatim}
{"column_name": "col",
 "cell_value": "value"}
\end{verbatim}

Similar to schema retrieval, cell retrieval uses query expansion to retrieve relevant column--value pairs from a retrieved table $t \in R_{q,k}$.
A language model $L$ generates up to five expanded queries $Q_{q,t}^{\mathrm{cell}}$ from the user query $q$.
Each expanded query $\hat{q} \in Q_{q,t}^{\mathrm{cell}}$ is encoded using $\mathcal{E}_{\mathrm{query}}$ to rank column--value pairs $p$ by the relevance score $r(\hat{q},p)$.
We take the union of the top-five pairs retrieved for each expanded query:
\begin{equation}
    V_{q,t}^{(b)}
    =
    \bigcup_{\hat{q} \in Q_{q,t}^{\mathrm{cell}}}
    \operatorname{Top5}_{p \in P_t^{(b)}}
    r(\hat{q},p),
\end{equation}
where $\operatorname{Top5}$ returns five pairs with the highest relevance scores.
The retrieved serialized JSON documents are provided to table reasoning.
The cell query expansion prompt is provided in Appendix~\ref{app:cell-query-expansion-prompts}.

\subsection{Table Reasoning}
\label{app:table-reasoning}

For each retrieved table $t \in R_{q,k}$, we construct a reasoning prompt containing the user query $q$, retrieved columns $S_{q,t}$, and retrieved column--value pairs $V_{q,t}^{(b)}$.
The retrieved column information includes data types and representative values, while column--value pairs are provided as serialized JSON strings.
The prompt is passed to a language model $L$ for table reasoning.
The prompt template is provided in Appendix~\ref{app:table-reasoning-prompt}.

In each ReAct iteration, $L$ generates a thought followed by an action or a final answer.
TableRAG~\citep{chen2024tablerag} executes Python actions through the PyReAct solver~\citep{liu-etal-2024-rethinking}, whereas our implementation uses SQL queries as actions instead.
SQL queries are executed through the HTTP API on our platform.
Query results and execution errors are returned to $L$ as the observations for subsequent iterations.
We use structured outputs to prevent parsing errors from interrupting the reasoning loop.

When $L$ produces an answer within $d$ iterations, the reasoning loop terminates, and a candidate answer is returned along with the corresponding SQL execution history.
If the iteration budget is exhausted without an answer, table reasoning returns an unanswerable outcome.

\subsection{Answer Aggregation}
\label{app:answer-aggregation}

\paragraph{SQL Similarity Ranking (SSR).}
For each candidate answer, we use the final executed SQL query as a document and encode the SQL query using $\mathcal{E}_{\mathrm{doc}}$.
The original user query $q$ is encoded using $\mathcal{E}_{\mathrm{query}}$.
We compute cosine similarity between the query embedding and each SQL query embedding to rank candidate answers.
The candidate answer with the highest similarity becomes the final answer.

\paragraph{LLM-based Answer Selection (LAS).}
We prompt a language model $L$ to perform answer selection.
The selection considers the table name of each candidate, final executed SQL, final thought, and answer, along with the user query $q$.
At most $k$ candidate answers are provided, along with an additional option to reject all candidates.
The answer selection prompt is provided in Appendix~\ref{app:answer-aggregation-prompt}.

\section{Answer Evaluation Rules}
\label{app:evaluation-rules}

We use hand-crafted rules to compare predicted answers with reference answers and report accuracy.
A predicted answer is considered correct if the applicable matching rule is satisfied, and incorrect otherwise.
The matching rules are divided into three categories based on the reference answer type: numerical, date, and text.

For numerical answers, we remove commas used as thousands separators and ignore unit suffixes in reference answers.
We extract numerical values from predictions and consider a match if any extracted value matches the reference value.
Numerical values are considered matching if values are equal or agree after rounding to the precision of the less precise value.
When the prediction is less precise than the reference answer, at least four significant digits are required.

For date answers, we accept equivalent date formats, including year-month-day and year-month representations.
Day-only predictions are also accepted for full-date reference answers.

For text answers, the reference answer must appear in the prediction.
For reference answers containing multiple comma-separated items, all items must match regardless of order.

When table reasoning fails to produce an answer from any retrieved table, the pipeline may return an empty prediction or an explicit no-answer response.
Both cases are counted as incorrect.

We do not perform semantic verification.
Predictions containing a correct value may be counted as correct even when the surrounding text does not answer the query.
Conversely, semantically equivalent answers may be counted as incorrect.

\clearpage
\onecolumn

\section{Prompts}
\label{app:prompts}

Figures~\ref{fig:prompt-schema-expansion}--\ref{fig:prompt-answer-aggregation} present illustrative prompts and example outputs for the main components of OpenTableRAG.
The examples use synthetic financial data.

\Needspace{8\baselineskip}
\subsection{Schema Query Expansion Prompt}
\label{app:schema-query-expansion-prompts}

\begin{tcolorbox}[
  breakable,
  left=1mm,
  right=1mm,
  top=1mm,
  bottom=1mm,
  colback=white
]
\begin{Verbatim}[breaklines=true,breakanywhere=true,fontsize=\scriptsize]
========================================= Prompt =========================================

Given a table of the financial database, I want to answer a question: What was the opening price of Apple Inc. (AAPL) on February 26, 2018?

Please suggest at most 5 column names or short column-name phrases that might contain the necessary data to answer this question. List the most important ones first.

Important:
- Prefer terms that are likely to appear directly in the table schema.
- Keep abbreviations, ticker symbols, currency codes, or technical terms from the question when they may appear directly in the schema, such as EPS, JPY, USD.
- Use concise column-like terms rather than full sentences.

Please answer in JSON format without any additional explanation.

Example:

{"keywords": ["Trade Date", "Company Name", "Close Price", "Volume"]}

======================================= Completion =======================================
{"keywords": ["Open Price", "Trade Date", "Ticker Symbol", "Company Name"]}
\end{Verbatim}
\end{tcolorbox}

\captionof{figure}{Illustrative example of the query expansion prompt for schema retrieval.}
\label{fig:prompt-schema-expansion}

\Needspace{8\baselineskip}
\subsection{Cell Query Expansion Prompt}
\label{app:cell-query-expansion-prompts}

\begin{tcolorbox}[
  breakable,
  left=1mm,
  right=1mm,
  top=1mm,
  bottom=1mm,
  colback=white
]
\begin{Verbatim}[breaklines=true,breakanywhere=true,fontsize=\scriptsize]
========================================= Prompt =========================================

Given a table of the financial database, I want to answer a question: What was the opening price of Apple Inc. (AAPL) on February 26, 2018?

Please extract at most 5 keywords which might appear in the table cells and help answer the question. List the most important ones first.

Important:
- The keywords should be contained in the question.
- The keywords should be categorical values, such as names or codes of stocks, funds, indices, currencies or categories (e.g. Microsoft, MSFT, SPY, JPY), rather than numerical values such as prices, amounts, ratios or dates.
- Keep the form used in the question.

Please answer in JSON format without any additional explanation.

Example:

{"keywords": ["Microsoft", "MSFT"]}

======================================= Completion =======================================
{"keywords": ["Apple Inc.", "AAPL"]}
\end{Verbatim}
\end{tcolorbox}

\captionof{figure}{Illustrative example of the query expansion prompt for cell retrieval.}
\label{fig:prompt-cell-expansion}

\Needspace{8\baselineskip}
\subsection{Table Reasoning Prompt}
\label{app:table-reasoning-prompt}

\begin{tcolorbox}[
  breakable,
  left=1mm,
  right=1mm,
  top=1mm,
  bottom=1mm,
  colback=white
]
\begin{Verbatim}[breaklines=true,breakanywhere=true,fontsize=\scriptsize]
========================================= Prompt =========================================

You answer a question with one table of the financial database, which you can only access by running SQL queries.

Question: What was the opening price of Apple Inc. (AAPL) on February 26, 2018?

Table: [DailyMarketIndicators]

Since you cannot view the table directly, here are some schemas and cell values retrieved from the table.

Schema Retrieval Queries: Open Price, Trade Date, Ticker Symbol, Company Name
Schema Retrieval Results:
{"column_name": "TradeDate", "dtype": "nvarchar(8)", "cell_examples": ["20180228", "20180227", "20180226"]}
{"column_name": "TickerSymbol", "dtype": "nvarchar(10)", "cell_examples": ["SPY", "QQQ", "IWM"]}
{"column_name": "CompanyName", "dtype": "nvarchar(100)", "cell_examples": ["SPDR S&P 500 ETF Trust", "Invesco QQQ Trust"]}
{"column_name": "OpenPrice", "dtype": "decimal(12,2)", "min": 0.01, "max": 9999.00}
...

Cell Retrieval Queries: Apple Inc., AAPL
Cell Retrieval Results:
{"column_name": "TickerSymbol", "cell_value": "SPY"}
{"column_name": "TickerSymbol", "cell_value": "QQQ"}
...

Database and SQL:
- The database is Microsoft SQL Server.
- Only read-only SELECT statements on [DailyMarketIndicators], one statement per step.
- Tables can hold hundreds of millions of rows, so always limit the data you read and return: filter with WHERE conditions, aggregate in SQL (COUNT, SUM, AVG, MIN, MAX, GROUP BY) and return only the rows you need.
- If a query returns an error, read the error, fix the SQL (syntax, table or column names) and run it again. If it times out or runs out of memory, split the computation into smaller queries or narrow the range.

At each step, think about what you know so far (thought), then either put one SQL query in action to run it (its result is shown at the next step), or put the final answer in answer and leave action empty. Give only the answer itself, without rounding.

Previous steps:
Thought: Search for the ticker across the table.
Action: SELECT TOP 10 [TickerSymbol], [OpenPrice] FROM [DailyMarketIndicators] WHERE [TickerSymbol] LIKE '%AAPL%'
Observation: Error: Execution timeout expired.

======================================= Completion =======================================
{"thought": "The broad wildcard search timed out. Restrict the query to the requested date to reduce the search scope.", "action": "SELECT TOP 10 [TickerSymbol], [OpenPrice] FROM [DailyMarketIndicators] WHERE [TradeDate] = '20180226' AND [TickerSymbol] LIKE '%AAPL%'", "answer": ""}

=================================== Next ReAct iteration ==================================
Previous steps:
...
Action: SELECT TOP 10 [TickerSymbol], [OpenPrice] FROM [DailyMarketIndicators] WHERE [TradeDate] = '20180226' AND [TickerSymbol] LIKE '%AAPL%'
Observation: Empty DataFrame

======================================= Completion =======================================
{"thought": "The revised SQL executed successfully, but the requested stock was not found in the selected table.", "action": "", "answer": "No data available"}
\end{Verbatim}
\end{tcolorbox}

\captionof{figure}{
Illustrative example of multi-round SQL querying adapted from an observed timeout recovery trace.
After a timeout, the SQL query is revised to restrict the search range and executes successfully.
The example uses a table other than the ground-truth table.
Some reasoning steps and retrieval results are omitted for space (``...'').
}
\label{fig:prompt-table-reasoning}

\Needspace{8\baselineskip}
\subsection{Answer Aggregation Prompt}
\label{app:answer-aggregation-prompt}

\begin{tcolorbox}[
  breakable,
  left=1mm,
  right=1mm,
  top=1mm,
  bottom=1mm,
  colback=white
]
\begin{Verbatim}[breaklines=true,breakanywhere=true,fontsize=\scriptsize]
========================================= Prompt =========================================

Several tables of the financial database were each queried with SQL to answer the same question. Choose the candidate whose answer is most likely correct, judging from whether its table, last SQL query and last thought match the question. If none of the candidates answers the question, choose 0.

Question: What was the opening price of Apple Inc. (AAPL) on February 26, 2018?

Candidates:
[1] table: PreMarketQuotes
last SQL: SELECT TOP 1 [TradeDate], [TickerSymbol], [CompanyName], [OpeningReferencePrice] FROM [PreMarketQuotes] WHERE [TradeDate] = '20180226' AND [TickerSymbol] = 'AAPL'
last thought: No matching record was found. The opening reference price is not necessarily the actual opening price.
answer: No opening price data available for Apple Inc. (AAPL) on February 26, 2018.

[2] table: StockOpeningReferences
last SQL: SELECT TOP 1 [TradeDate], [TickerSymbol], [OpeningReferencePrice] FROM [StockOpeningReferences] WHERE [TradeDate] = '20180226' AND [TickerSymbol] = 'AAPL'
last thought: The opening reference price is 44.50, but this value does not represent the actual opening price.
answer: The opening reference price was 44.50 USD, which may differ from the actual opening price.

[3] table: OptionVolatilityData
last SQL: SELECT TOP 1 [OpenPrice] FROM [OptionVolatilityData] WHERE [TradeDate] = '20180226' AND [TickerSymbol] = 'AAPL'
last thought: No matching record was found. This table contains option volatility data rather than daily stock prices.
answer: No opening price data available in this table.

[4] table: DailyStockPrices
last SQL: SELECT [OpenPrice] FROM [DailyStockPrices] WHERE [TradeDate] = '20180226' AND [TickerSymbol] = 'AAPL'
last thought: Retrieve the opening price using the specified trade date and ticker symbol.
answer: 45.20

[5] table: WeeklyStockPrices
last SQL: SELECT [OpenPrice] FROM [WeeklyStockPrices] WHERE [TradeDate] = '20180226' AND [TickerSymbol] = 'AAPL'
last thought: No matching weekly record was found for the specified date.
answer: No opening price data available for the specified date.

...

======================================= Completion =======================================
{"thought": "Candidate 4 uses daily stock prices and retrieves the opening price for the specified trade date and ticker symbol, matching the question.", "choice": 4}
\end{Verbatim}
\end{tcolorbox}

\captionof{figure}{
Illustrative example of answer aggregation with LLM-based answer selection (LAS).
Additional candidates are omitted for space (``...'').
}
\label{fig:prompt-answer-aggregation}

\clearpage
\twocolumn

\end{document}